Authors, affiliations and addresses:

**Rodrigo Arnaldo Scarpel,** Instituto Tecnológico de Aeronáutica, Praça Marechal Eduardo Gomes, 50 ITA - IEM sala 2311, São José dos Campos - SP, Brazil, rodrigo@ita.br

**Alexandros Ladas,** University of Nottingham, School of Computer Science, Jubilee Campus, Wollaton Road, Nottingham, UK, NG8 1BB, psxal2@nottingham.ac.uk

**Uwe Aickelin,** University of Nottingham, School of Computer Science, Jubilee Campus, Wollaton Road, Nottingham, UK, NG8 1BB, uwe.aickelin@nottingham.ac.uk

Correspondig author: Rodrigo Arnaldo Scarpel, rodrigo@ita.br, +55(12)3947-6973

Title: Indebted households profiling: a knowledge discovery from database approach

Abstract:
A major challenge in consumer credit risk portfolio management is to classify households according to their risk profile. In order to build such risk profiles it is necessary to employ an approach that analyses data systematically in order to detect important relationships, interactions, dependencies and associations amongst the available continuous and categorical variables altogether and accurately generate profiles of most interesting household segments according to their credit risk. The objective of this work is to employ a knowledge discovery from database process to identify groups of indebted households and describe their profiles using a database collected by the Consumer Credit Counselling Service (CCCS) in the UK. Employing a framework that allows the usage of both categorical and continuous data altogether to find hidden structures in unlabelled data it was established the ideal number of clusters and such clusters were described in order to identify the households who exhibit a high propensity of excessive debt levels.

Keywords: Clustering, Homogeneity analysis, Silhouette width, credit risk.

**Indebted households profiling: a knowledge discovery from database approach**

# 1. Introduction

Indebtedness in private households as a result of growing consumer credit use has dramatically risen and according to Kamleitner and Kirchler [1] it has various consequences on social, psychological, economic, and political levels. As reported by McCarthy [2], financial distress at an individual and household level can have serious consequences which go far beyond those experienced by the individual or household involved. Furthermore, the enormous fiscal costs associated with a financial crisis are a reminder that heightened financial distress and poor financial behaviour on the part of a relatively small number of people can have serious negative externality effects on the rest of the economy.

In the literature, there are different works that employed data mining approaches to deal with credit risk assessment. Shi et al. [3] obtained promising results on bankruptcy prediction employing a multiple criteria linear programming (MCLP) approach to data mining. Peng et al. [4] employed cluster analysis for credit card accounts classification and improved clustering classification results using ensemble and supervised learning methods. Aihua et al. [5] proposed a data mining approach based on the combination of Multi-criteria linear programming and Principal Component Analysis in order to improve the classification of credit cardholders. Peng et al. [6] proposed a mathematical programming model to deal with Credit Classification Problems addressing speed and scalability that are two essential issues in data mining and knowledge discovery. Li, Shi and He [7] proposed three multiple criteria linear programming (MCLP) to improve the overall accuracy of the classification models. The first one is called MCLP with unbalanced training set selection, the second one is called fuzzy linear programming (FLP) method with moving boundary, and the third one is called penalized multi criteria linear programming (PMCLP). This work intends to contribute to the existing literature and to consumer credit risk portfolio management by providing an approach to classify households according to their risk profile. Such profiling is useful in different ways. On the aggregate level is important to model the distribution of the expected number of defaults according to both the economic factors and the social and demographic developments, such as the increase of the number of divorces. Moreover, events like recessions may impact some households more acutely than others. Therefore, the identified profiles are useful in indicating which socio-economic factors should be considered on such macro level models. Households' profiles may also support the development of a portfolio level forecast by clustering the individuals according to the generated profiles. Thus, one needs only to sum across predictions by cluster in order to produce a portfolio level forecast.

In order to build such risk profiles it is necessary to employ an approach that analyses data systematically in order to detect important relationships, interactions, dependencies and associations amongst the available continuous and categorical variables altogether and accurately generate profiles of most interesting household segments according to their credit risk. This way, the objective of this work is to employ a knowledge discovery from database (KDD) process to identify groups of households and describe their profiles using a database collected by the Consumer Credit Counselling Service (CCCS) in the UK. Thus, it was proposed a framework to meet two requirements: First, it must allow the usage of both categorical and continuous data altogether; second, it must be unsupervised, since we are trying to find hidden structures in unlabelled data. According to Disney and Gathergood [8], the main advantage of using the CCCS data for analytical purposes is that the counselling agency collects a wide range of socio-economics and financial data at the individual level, providing a complete picture of client assets and debts. Furthermore, clients of counselling organizations have an incentive to reveal true information to debt counsellors in order to gain better financial advice.

The decision to employ a KDD process in this study is based on the observation that CCCS has collected high volumes of data related to different aspects of the interaction that takes place between demographics, expenditure and debt. KDD is defined as the nontrivial process of identifying valid, novel, potentially useful, and ultimately understandable patterns in data [9] and it is often used to describe the process of extracting and identifying patterns from a large amount of available data. Data mining is a particular step of the KDD process and it includes selecting methods to be used for searching for patterns in the data, as well as searching for patterns of interest in a particular representation form or a set of such representations [9,10].

The rest of the paper is organized as follows. In Section 2 the employed methodology is described. Section 3 has three sub-sections, the first focuses on the data selection, the second on the pre-processing and transformation steps and the third focuses on the profiles building and interpretation. Section 4 concludes the paper.

**2. Methodology**

This study presents a framework based on the KDD process to identify groups of households and describe their profiles using a database collected by the Consumer Credit Counselling Service (CCCS) in the UK. The KDD process is interactive and iterative involving numerous steps, summarized as [10,11]:

1. Data selection: includes selecting some datasets or focusing on a subset of variables on which the analysis will be performed;
2. Data pre-processing: includes data cleaning and other operations, such as removing duplicate data and outliers, as well as deciding on strategies for handling missing data fields;
3. Data transformation: includes finding useful features to represent the data, depending on the goal of the task, and using dimensionality reduction or transformation methods to reduce the effective number of variables under consideration or find invariant representation for the data;
4. Data mining: includes selecting methods to be used for searching for patterns in the data, as well as searching for patterns of interest in a particular representation form or a set of such representations;
5. Interpretation: includes interpreting the discovered patterns, generating possible visualization of the extracted patterns and using the discovered knowledge (incorporating it into a performance system, taking actions based on it or simply documenting and reporting it to interested parties).

As mentioned before, the framework used in this study needed to meet two requirements: First, it must allow the usage of both categorical and continuous data together; second, it must be unsupervised, since we are trying to find hidden structures in unlabelled data. Figure 1 shows the employed framework. It follows a sequential procedure composed by the steps: (1) data selection; (2) pre-processing and data transformation; and (3) profiles building and interpretation. Concerning the usage of both continuous and categorical data together, on the pre-processing and data transformation step, we chose to categorize the continuous data and submit both the categorical and categorized data to the data reduction procedure. For such data reduction the homogeneity analysis is employed not only because it is appropriate to categorical data but also due its capacity to perform an optimal scaling to estimate the coordinates of the observations in a reduced space. On the profiles building and interpretation step a data clustering procedure is applied taking the estimated coordinates as input and the obtained clusters are described in order to build the intended profiles.

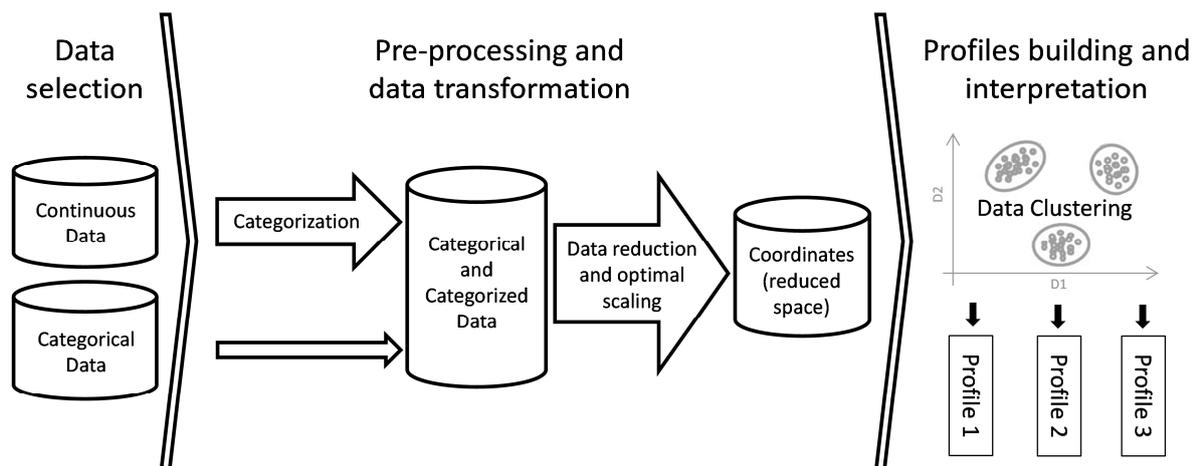

Figure 1: Employed framework

2.1 Data reduction and optimal scaling:

As mentioned before, the homogeneity analysis was employed for the data reduction and optimal scaling task. In homogeneity analysis we try to find a low-dimension space in which observations and categories are positioned in such a way that as much information as possible is retained from the original data [12]. Hence, the goal is to construct a low-dimensional joint map of observations and categories in Euclidean space ($\Re^p$), where p is the number of dimensions on the reduced space. The homogeneity analysis can be stated as follows: consider a data matrix $H$ (N × J) in which the rows correspond to N observations measured on J categorical or categorized variables, where j (=1,…,J) can take $l_j$ possible values (categories). Let k = ($k_l$, . . . , $k_j$, . . . , $k_J$) be the J-vector containing the number of categories of each variable and let each variable $h_j$ (j = 1,…,J) be coded into an (N x $k_j$) indicator matrix $G_j$. Define $X$ as a (N × p) matrix of observations scores and define J ($k_j$ x p) matrices $Y_j$ of category quantifications. Homogeneity analysis then amounts to minimizing

$$J^{-1} \sum_{j=1}^{J} \mathrm{tr}\left[(X - G_j Y_j)'(X - G_j Y_j)\right]$$

simultaneously over $X$ and $Y_j$'s under the restrictions

$$X'X = NI_p$$

$$1'X = 0$$

Such restriction are necessary in order to avoid the trivial solution, corresponding to $X = 0$ and $Y_j = 0$ (j=1,…,J). The first restriction standardizes the squared length of the observation scores (to be equal to N), and in two or higher dimensions also requires the columns of $X$ to be orthogonal. The second restriction is for normalization and to generate a centred around the origin graph plot.

By imposing other restrictions on the category quantifications $Y_j$, and in some cases on the coding of the data, different types of analysis can be derived. In this work, as we are employing both categorical and categorized data together, an additional restriction is required to make the categorized data work as ordinal data. Thus, we imposed that category quantifications $Y_j$ are estimated using

$$Y_j = o_j \beta_j', \quad j = 1,...,J$$

where $o_j$ is a $l_j$ column vector of single ordinal quantifications for variable j and $\beta_j$ is a p-column vector of weights. By imposing this new restriction the decision variables of the problem become both the matrix of observations scores $X$ and the vector of weights $\beta_j$ and the problem is solved employing the Alternating Least Squares (ALS) method proposed by Michailidis and de Leeuw [12]. In order to illustrate the effect of such model modification, in Figure 2 it is shown the results we obtained by applying both the standard homogeneity analysis and the homogeneity analysis with the additional restriction on a 7 observations and 2 ordinal variables example. Comparing the results (Figure 2) it is possible to observe that the model with the additional restriction was able to incorporate into the analysis the underlying monotone structure in the data and thus have treated the variables as ordinal.

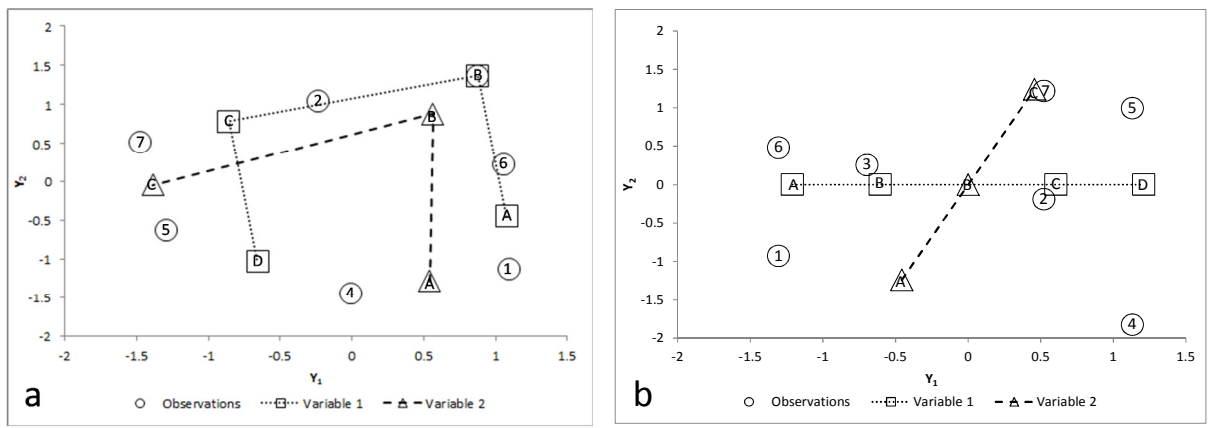

Figure 2: Homogeneity analysis results: (a) standard model; (b) model with the additional restriction.

2.2 Clustering:

As one of the major data mining functionalities, clustering has broad application in different areas such as medicine, psychology, sociology, food and sensory research, business economics, marketing research and many others. Cluster analysis has been known under different names, according to the discipline, such as numerical taxonomy, Q-analysis, unsupervised pattern recognition, clumping and group analysis [13].

According to Webb [14], clustering analysis is the grouping of individuals of a population in order to discover structures in the data. Ideally, one would like the observations within a group to be close or similar to one another, but dissimilar from observations in other groups. The problem addressed by cluster analysis can be stated as follows: Given a collection of $n$ observations, each of which is described by a set of $k$ variables, derive a partition into a number of clusters that are internally homogeneous and externally heterogeneous.

Three major categories of cluster analysis can be distinguished: non-overlapping methods, overlapping methods and fuzzy methods. In the case of non-overlapping methods an entity belongs to one and only one cluster. The assumption of external isolation is relaxed in methods providing overlapping and fuzzy partitioning. In overlapping clusters, an entity may belong to more than one cluster. In the case of fuzzy clusters, entities have partial membership in more than one cluster [13]. There is a vast literature on clustering methods and for a complete review see Duda, Hart and Stork [15], Webb [14] and Liao [16].

Identifying the appropriate clustering technique and the ideal number of clusters are two crucial issues in clustering. According to Saha and Bandyopadhyay [17], model selection in clustering consists of two steps. In the first step the proper clustering method for a particular data set has to be identified. Once this choice is made, the model order remains to be determined for the given data set in the second step. The task of determining the number of clusters and also the validity of the clusters formed are generally addressed by providing several definitions of validity indices. The measure of validity of clusters should be such that it will be able to impose an ordering of the clusters in terms of its goodness. A commonly used index for the model selection and the ideal number of clusters identification is the silhouette width [18]. Such index combines measures of compactness and separation of the clusters. The silhouette width is measured by $[b(i) - a(i)]/\max[a(i),b(i)]$, where $a(i)$ is the average dissimilarity between observation i and all other points of the cluster to which i belongs and $b(i)$ is the smallest average dissimilarity of i to all observations of all other clusters. The silhouette width value measures the degree of confidence in a particular clustering assignment and lies in the interval [-1,1], with well-clustered observations having values near 1 and poorly clustered observations having values near -1.

In this work the sequence suggested by Saha and Bandyopadhyay [17] is applied considering the K-means, partitioning around medoids (PAM), clustering large applications (CLARA) and fuzzy analysis clustering (FANNY) as partitioning techniques candidates and the silhouette width as validity index.

**3. Analysis and results**

In order to identify groups of indebted households, the framework described in Section 2 is applied to an individual level dataset of clients in the UK provided by the Consumer Credit Counselling Service (CCCS). CCCS collect data on client assets and debt, income and expenditures and demographic information. It was provided data on all clients who contacted the organization between January 2004 and December 2008 totalizing about 75,000 clients. Data was provided in anonymised format with all individual identifiers removed and replaced with a single randomly-assigned 8-digit numerical identifier [8].

3.1 Data selection:

On the data selection step a set of attributes were selected among the available data. For the continuous data, because the absolute amount of debt and expenditure is cumulative, it cannot represent the level, intensity of severity of debt and expenditure. Thus, instead of using the absolute amount it was used the debt-to-income and spending-to-income ratios. Table 1 lists the variables included in this study, as well as their description.

3.2 Pre-processing and data transformation:

At the pre-processing stage it is essential to test the quality of the collected data and to filter out information with no significance to our study. Since there were repeated observations for an individual and individuals with no information on spending in none of the types, such duplicated and missing data were removed. After that, the sample reduced to about 41,000 clients. We also evaluated the continuous variables concerning their symmetry, skewness as well as the presence of outliers. Figure 3 shows the boxplots of the employed continuous variables.

From Figure 3, it is possible to notice that all continuous variables present a highly skewed distribution with many extreme values. The usage of such data must be done with extreme careful since they can seriously bias or influence estimates. Thus, instead of performing some transformation to normalize the distributions and remove the extreme values, on the data transformation stage we have chosen to categorize the continuous data. Therefore, such variables were categorized into different number of ordinal categories. Table 2 shows the categorized variables and their ordinal categories.

Table 1: Description of the employed data

| Variable | Description |
|---|---|
| Mstat – Depen | Marital status - dependents (couple-cohabiting - with dependents, couple-cohabiting - no dependents, married - no dependents, married - with dependents, divorced - no dependents, divorced - with dependents, separated - no dependents, separated - with dependents, single - no dependents, single - with dependents, other - no dependents, other - with dependents) |
| Empstat | Employment status (retired, unemployed, self-employed, employee-full time, employee-part time, other) |
| Hstatus | Housing ownership status (home owner with mortgage, outright owner, renter) |
| ClothingtInc | Spending on clothing to income |
| FoodtInc | Spending on food to income |
| ServicestInc | Spending on services to income |
| HousingtInc | Spending on housing to income |
| MotoringiInc | Spending on motoring to income |
| LeisuretInc | Spending on leisure to income |
| TDebttInc | Total debt to annual income |

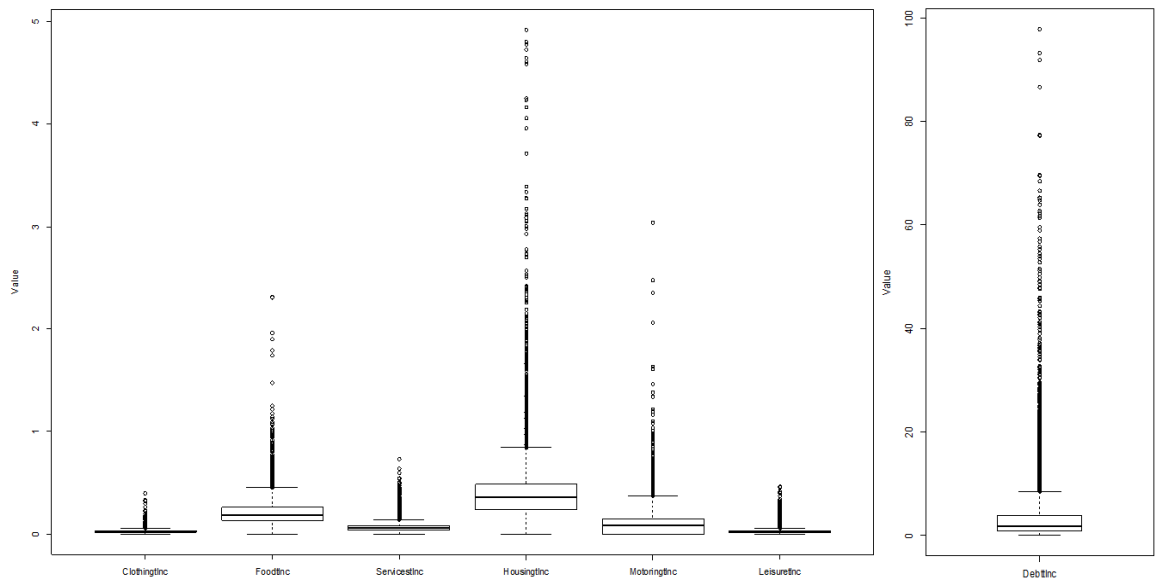

Figure 3. Boxplots of the continuous variables.

The next step into the pre-processing and data transformation stage is the data reduction and optimal scaling. To perform such task it was employed the homogeneity analysis with an additional restriction that allows considering the underlying monotone structure in the ordinal data, as described before. The result of the homogeneity analysis is a low-dimensional joint map of objects and categories in Euclidean space ($\Re^p$). For this study we have chosen to work in the 2-dimensions space (p=2) not only because it makes the interpretation of the structural relationship between the used socio-economics and financial data simpler but also to support the validation of the identified profiles. Figures 4 and 5 show the homogeneity analysis results. Figure 4 shows both the object scores (coordinates of the observations) and the coordinates of the ordinal variables (category points) in the reduced space (p=2). The category points were represented using arrows to indicate the ordinal increasing direction. Figure 5 shows the coordinates of the categorical data (category points) in the reduced space (p=2).

Table 2: Categorized variables and generated ordinal categories.

| Variable | Ordinal categories |
|---|---|
| Clothing | 1: from 0 to 0.0125, 2: from 0.0125 to 0.0175, 3: from 0.0176 to 0.020, 4: from 0.021 to 0.025, 5: from 0.026 to 0.030, 6: from 0.031 to 0.040, 7: from 0.041 to 0.050, 8: from 0.051 to 0.075, 9: higher than 0.076 |
| Food | 1: from 0 to 0.100, 2: from 0.101 to 0.125, 3: from 0.126 to 0.150, 4: from 0.151 to 0.175, 5: from 0.176 to 0.200, 6: from 0.201 to 0.250, 7: from 0.251 to 0.333, 8: from 0.334 to 0.500, 9: higher than 0.501 |
| Services | 1: from 0 to 0.030, 2: from 0.031 to 0.040, 3: from 0.041 to 0.050, 4: from 0.051 to 0.060, 5: from 0.061 to 0.070, 6: from 0.071 to 0.085, 7: from 0.086 to 0.100, 8: from 0.101 to 0.150, 9: higher than 0.151 |
| Housing | 1: from 0 to 0.150, 2: from 0.151 to 0.250, 3: from 0.251 to 0.333, 4: from 0.334 to 0.400, 5: from 0.401 to 0.500, 6: from 0.501 to 0.600, 7: from 0.601 to 0.750, 8: higher than 0.751 |
| Motoring | 1: from 0 to 0.001, 2: from 0.002 to 0.050, 3: from 0.051 to 0.100, 4: from 0.101 to 0.150, 5: from 0.151 to 0.200, 6: from 0.201 to 0.300, 7: higher than 0.301 |
| Leisure | 1: from 0 to 0.010, 2: from 0.011 to 0.015, 3: from 0.016 to 0.020, 4: from 0.021 to 0.025, 5: from 0.026 to 0.035, 6: from 0.036 to 0.050, 7: higher than 0.051 |
| Total Debt | 1: from 0 to 0.333, 2: from 0.334 to 0.50, 3: from 0.51 to 0.75, 4: from 0.76 to 1.00, 5: from 1.01 to 1.25, 6: from 1.26 to 1.50, 7: from 1.51 to 2.00, 8: from 2.01 to 3.0, 9: from 3.1 to 5.0, 10: from 5.1 to 7.5, 11: from 7.6 to 10.0, 12: from 10.1 to 15.0, 13: higher than 15.1 |

From Figure 4 it is possible to see that the first dimension is a measure of overall spending-to-income rate, since all spending type exhibit high dispersion through it. However, it is interesting to note that not all spending types are positively highly correlated. There is a high linear relationship between the spending on food, clothes and services, but such spending types are negatively correlated to spending on both housing and motoring, that are highly correlated to each other. The spending on leisure is positively correlated to the spending on food, clothing and services, but it is not possible to indicate that their relationship is high. Concerning the total level of debt, it is positively correlated to the first dimension and negatively correlated to the second dimension, but it is not possible to say that it is highly correlated to any of the spending types. Thus, later on the profiles identification step, we expect to identify clusters of households with different spending patterns having problem with different debt types.

3.3 Profiles building and interpretation:

The next step is the profiles building. Since we are trying to find hidden structures in unlabelled data, we have chosen to perform clustering employing partitioning techniques. Thus, the sequence suggested by Saha and Bandyopadhyay [17] was applied considering the K-means, PAM, CLARA and FANNY as the partitioning techniques candidates and the silhouette width as validity index in order to determine both the more appropriate partitioning technique and the ideal number of clusters. Figure 6 presents the estimated silhouette width ranging the number of clusters from 2 to 8. There are no silhouette width values for the FANNY when the number of clusters is 7 and 8 because in such cases the method did not converge.

From Figure 6, it is possible to see that the ideal number of clusters is 3 since it is the value with the maximum silhouette width for all the partitioning techniques. Concerning the partitioning technique selection, it is possible to see that the estimated silhouette width value for the K-means is slightly superior then the estimated value for PAM, CLARA and FANNY. Thus, we have chosen to build the clusters of households employing the K-means partitioning technique. In order to describe the obtained clusters according to the financial distress that such households are facing we additionally used the debt-to-income ratio by debt type besides that data used to build the clusters. Such debt-to-income ratios give an indication of the borrower's ability to repay the debt after it is disbursed. Table 3 gives the mean values of the continuous variables for each cluster and Table 4 gives the relative frequency of the categorical variables through the clusters, as well as their lift. The lift value is obtained dividing the relative frequencies of the variables by the relative frequency of the observations through the clusters. Such value is used to compare the frequency of a category against a baseline frequency computed under the statistical independence assumption. Thus, a lift value near 1 indicates a weak correlation between the category and the cluster, values higher than 1 indicates a positive correlation between the category and the cluster and values lower than 1 indicates a negative correlation between the category and the cluster.

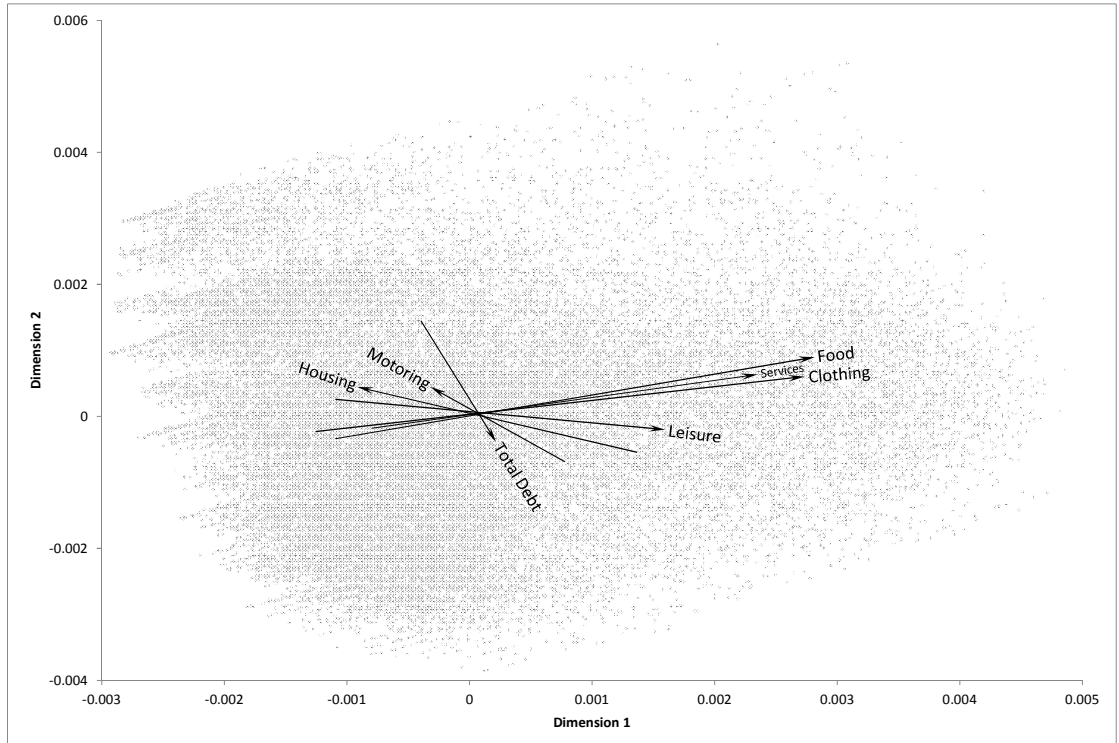

Figure 4. Objects score and the ordinal variables category points in the reduced space (p=2).

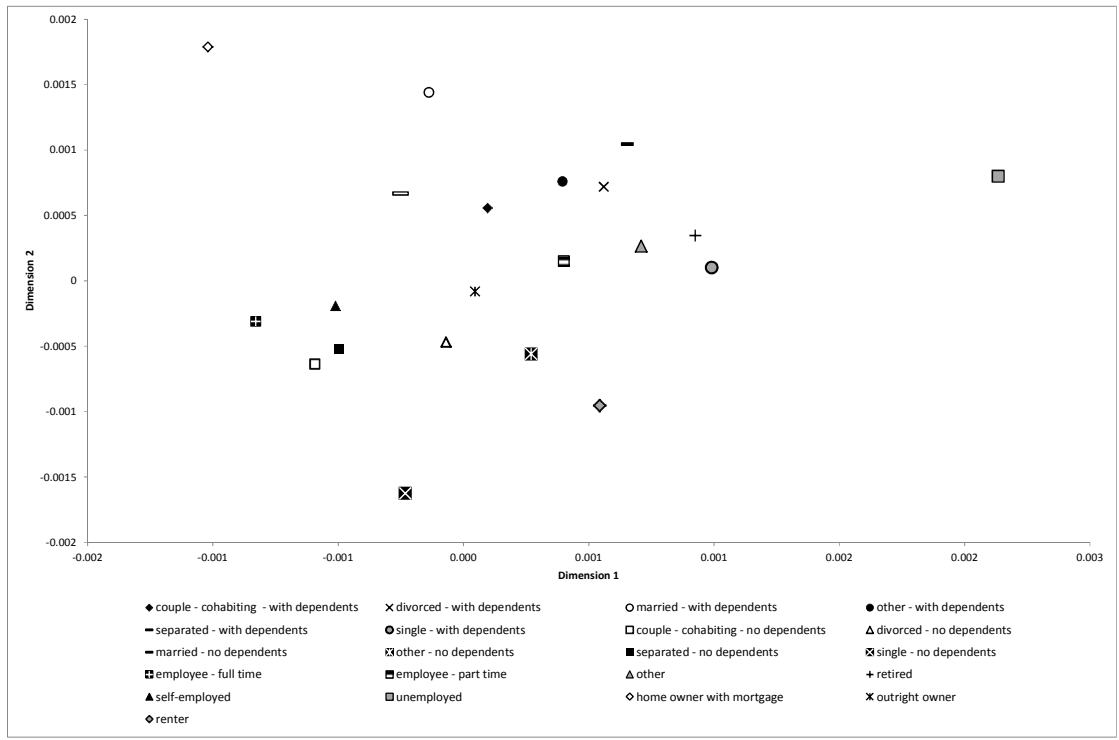

Figure 5. Coordinates of the categorical variables in the reduced space (p=2).

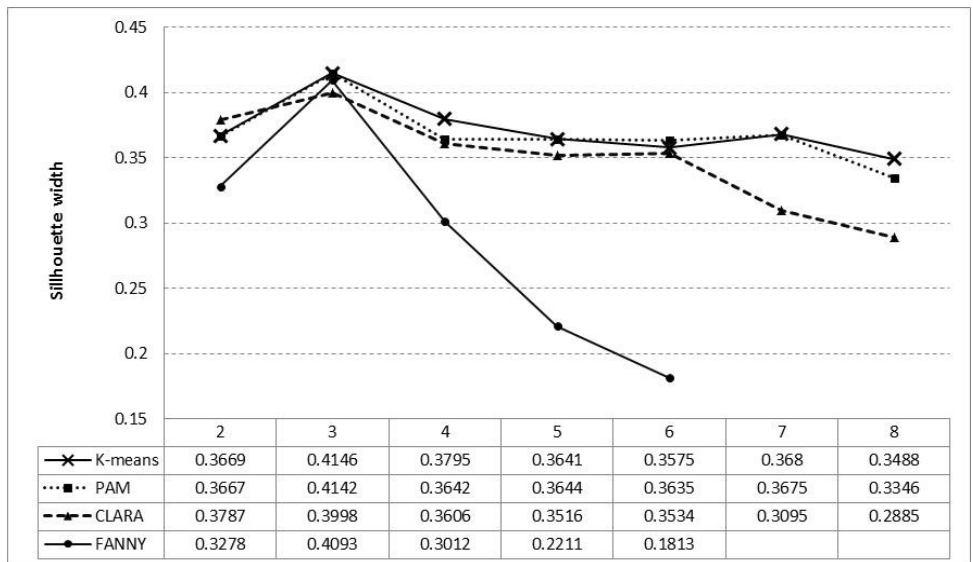

Figure 6. Obtained silhouette width ranging the number of clusters from 2 to 8 for the K-means, partitioning around medoids (PAM), clustering large applications (CLARA) and fuzzy analysis (FANNY) partitioning techniques.

Table 3: Mean values by continuous variable and cluster

|  | Variable | Cluster | | | Total |
|---|---|---|---|---|---|
|  |  | 1 | 2 | 3 |  |
| Spending type | Clothing | 0,021 | 0,021 | **0,044** | 0,027 |
|  | Food | 0,175 | 0,160 | **0,335** | 0,210 |
|  | Services | 0,058 | 0,055 | **0,098** | 0,067 |
|  | Leisure | 0,017 | 0,024 | **0,040** | 0,026 |
|  | Housing | **0,479** | 0,360 | 0,281 | 0,379 |
|  | Motoring | **0,109** | 0,099 | 0,081 | 0,098 |
| Debt type | Mortgage | **3,228** | 0,267 | 0,733 | 1,372 |
|  | Catalogues | 0,016 | 0,018 | **0,058** | 0,028 |
|  | Collection agency | 0,064 | 0,078 | **0,103** | 0,080 |
|  | Credit card | **0,774** | 0,570 | 0,699 | 0,671 |
|  | Personal Loan | 0,681 | **0,800** | 0,745 | 0,746 |
|  | Store Card | 0,016 | 0,014 | **0,024** | 0,017 |

From Tables 3 and 4 it is possible to see that 33.2% of the households were assigned to the first cluster. Concerning the marital and employment status such cluster is essentially composed by married people with dependents or couples cohabiting with dependents and most of them are home owners with mortgages. In terms of employment status, there is a high concentration of self-employed people. About their spending pattern is possible to say that it is highly concentrated in housing and motoring and such households have a huge debt on mortgage and credit card. Concerning this first cluster, it is expected that households with children have a higher level of debt because of the high household expenses associated with the children. Cunha, Lambrecht and Pawlina [19] studied the determinants and behaviour of outstanding mortgage loan-to-value (LTV). The LTV is one of the most significant measures of credit risk on a mortgage. According to such authors, the number of household members is shown to positively affect LTV. Moreover, according to Georgarakos, Lojschova and Ward-Warmedinger [20], households may have not fully taken into account the costs associated with children before deciding to borrow and thus children show up as an additional source of distress when servicing mortgage debt.

Another source of distress for the first cluster is associated with the uncertainty of such households' income since there is a high concentration of self-employed people. In a study of the determinants of mortgage delinquency across 12 EU countries Diaz-Serrano [21] documented a positive association between income volatility and the risk of mortgage delinquency. In summary, it is possible to indicate that the households in the first cluster are facing financial distress due to both their family size and their income uncertainty.

Table 4: Relative frequency and lift of the categorical variables by cluster

| | Variable | Relative frequency | | | Lift | | |
|---|---|---|---|---|---|---|---|
| | | Cluster | | | Cluster | | |
| | | 1 | 2 | 3 | 1 | 2 | 3 |
| Marital status - dependents | couple - cohabiting - no dependents | 25.6% | 63.8% | 10.6% | 0.770 | **1.566** | 0.407 |
| | couple - cohabiting - with dependents | 41.5% | 27.9% | 30.6% | **1.248** | 0.685 | 1.177 |
| | divorced - no dependents | 20.0% | 53.5% | 26.6% | 0.601 | **1.311** | 1.021 |
| | divorced - with dependents | 36.8% | 21.0% | 42.2% | 1.107 | 0.514 | **1.625** |
| | married - no dependents | 48.4% | 26.3% | 25.3% | **1.457** | 0.646 | 0.972 |
| | married - with dependents | 66.1% | 7.2% | 26.7% | **1.989** | 0.177 | 1.027 |
| | other - no dependents | 17.9% | 51.6% | 30.5% | 0.538 | **1.266** | 1.173 |
| | other - with dependents | 44.0% | 17.9% | 38.1% | 1.324 | 0.439 | **1.466** |
| | separated - no dependents | 25.6% | 57.2% | 17.2% | 0.770 | **1.403** | 0.663 |
| | separated - with dependents | 42.4% | 13.5% | 44.1% | 1.278 | 0.331 | **1.694** |
| | single - no dependents | 7.5% | 79.1% | 13.3% | 0.226 | **1.941** | 0.513 |
| | single - with dependents | 18.7% | 33.4% | 47.9% | 0.564 | 0.819 | **1.841** |
| Employment status | employee - full time | 36.7% | 57.3% | 6.0% | 1.106 | **1.404** | 0.230 |
| | employee - part time | 38.7% | 28.0% | 33.3% | 1.166 | 0.687 | **1.279** |
| | other | 28.4% | 27.7% | 43.9% | 0.856 | 0.679 | **1.688** |
| | retired | 20.1% | 26.0% | 53.9% | 0.606 | 0.636 | **2.074** |
| | self-employed | 54.2% | 34.1% | 11.6% | **1.633** | 0.837 | 0.447 |
| | unemployed | 15.6% | 6.7% | 77.6% | 0.471 | 0.165 | **2.985** |
| Housing status | home owner with mortgage | 85.2% | 5.6% | 9.2% | **2.565** | 0.137 | 0.355 |
| | outright owner | 30.7% | 42.9% | 26.4% | 0.923 | 1.053 | 1.015 |
| | renter | 5.8% | 58.7% | 35.5% | 0.174 | **1.438** | 1.367 |
| | Total | 33.2% | 40.8% | 26.0% | | | |

From Tables 3 and 4 it is possible to see that 40.8% of the households were assigned to the second cluster. It is the biggest generated cluster and it composed by people with different marital status but in all cases with no dependents. Concerning their employment and housing status, there is a high concentration of full time employees living in rented houses. The spending pattern of such households is near the mean value in all spending types and in terms of debt it is possible to indicate that the only debt-to-annual income ratio higher than the mean value is on personal loans. Such cluster also presents the lowest mean value for the total debt-to-annual income ratio of such households (1.87 for cluster 2, 4.91 for cluster 1 and 2.51 for cluster 3). According to Mann, Mann and Staples [22], debt may arise for any of several reasons: as a device to smooth consumption over the lifetime, as a response to some previous shock (medical problems, job loss or any unexpected event), or because of a failure to manage consumption to match income. It is not possible to indicate how come the households of cluster 2 required or perceived themselves to require counselling, however, since they aren't associated with any of the known sources of financial distress, it seems that they will be able to pay for their debt. In summary, the households of the second cluster are probably facing financial distress due to the usage of credit to obtain funds to mitigate the adverse effects of the happened event.

Concerning cluster 3, Tables 3 and 4 show that 26.0% of the households were assigned to this cluster. It is the smallest generated cluster and it composed by people with low level of participation in the labour force since there is a high concentration of unemployed, retired and part-time employees. Such households present different marital status (divorced, separated, single and other), but in all cases they have dependents. About their spending behaviour, there is high concentration on clothing, food, services and leisure and their debt is highly concentrated on catalogues, collection agencies and store cards. The part-time employees can be described as working-class households, earning less for their work and working despite the presence of young children. According to Salomon and Ben-Akiva [23], such people, in reference to the lifestyle choice, includes those households who have chosen to establish a family with children and who have chosen, in most cases by default, to participate in lower paying jobs in the labour market. Another sources of distress for the third cluster are job loss and the financial complications associated with divorces [24]. According to Keys [25], the effects of job loss are particularly complex and even relatively brief unemployment spells can have significant long-term consequences on households' credit market outcomes. In summary, it is possible to indicate that due to problems associated to the available income the households of the third cluster are incurring to multiple debts and are exposed to financial distress in order to provide for themselves and their families.

## 4. Conclusions

In order to contribute to both the existing literature on credit risk assessment and to consumer credit risk portfolio management, this work provided an approach to classify households according to their risk profile. Therefore, a KDD process was employed to identify groups of households and describe their profiles, concerning their socio-economic characteristics, using a database collected by the CCCS in the UK. Such profiling is useful to model the distribution of the expected number of defaults according to both the economic factors and the social and demographic developments and to indicate which socio-economic factors should be considered on such models. Households' profiles may also support the development of a portfolio level forecast by clustering the individuals according to the generated profiles.

By using the proposed framework data were analysed systematically to find hidden structures in unlabelled data. Such framework also allowed the usage of both categorical and continuous data altogether. It was established the ideal number of clusters and such clusters were described in order to identify the households who exhibit a high propensity of excessive debt levels. It was also possible to describe the obtained clusters according to the financial distress that such households are facing.

The main drawback of using the CCCS data is that it was collected from a group of individuals who require (or perceive themselves to require) counselling. Thus, the scope of this work was limited to the identification of patterns within a particular sub-group of the population. For future work we intend gather further data not only to validate the identified profiles but also to build a complete picture of consumer indebtedness.

**References**

[1] Kamleitner B, Kirchler E (2007) Consumer Credit Use: A Process Model and Literature Review. European Review of Applied Psychology, 57 (4), 267–283.
[2] McCarthy I (2011) Behavioural characteristics and financial distress. Working Paper Series 1303, European Central Bank.
[3] Shi Y, Peng Y, Xu W, Tang X (2002) Data Mining via Multiple Criteria Linear Programming: Applications in Credit Card Portfolio Management. International Journal of Information Technology and Decision Making, 1, 131–151.
[4] Peng Y, Kou G, Shi Y, Chen Z (2005) Improving Clustering Analysis for Credit Card Accounts Classification. In V. S. Sunderam et al, eds., ICCS 2005, LNCS 3516, Springer, Berlin, 548-553.
[5] Aihua L, Shi Y, Zhu M, Dai J (2006) A Data Mining Approach to Classify Credit Cardholders' Behavior. Proceeding of workshops on The Sixth IEEE International Conference on Data Mining (ICDM), HongKong, Dec 19-22.
[6] Peng Y, Kou G, Shi Y, Chen Z (2008) A Multi-Criteria Convex Quadratic Programming Model for Credit Data Analysis. Decision Support Systems, 44, 1016-1030.
[7] Li A, Shi Y, He J (2008) MCLP-based Methods for Improving "Bad" Catching Rate in Credit Cardholder Behavior Analysis. Applied Soft Computing, 8(3), 1259-1265.
[8] Disney R, Gathergood J (2009) Understanding consumer over-indebtedness using counselling sector data: Scoping study. Report to the Department for Business, Innovation and Skills (BIS).
[9] Frawley W J, Platetsky-Shapiro G, Matheus C J (1991) Knowledge Discovery in databases: An Overview. In G. Ratetsky-Shapiro and B. Frawley (Ed.), Knowledge Discovery in databases, Cambridge, Mass: AAAI/MIT Press, pp. 1–27, 1991.
[10] Han J, Kamber M (2001) Data Mining: Concepts and Techiniques, 1st edition. New York: Morgan Kaufmann.
[11] Fayyad U M, Piatetsky-Shapiro G, Smyth P (1996) The KDD process for extracting useful knowledge from volumes of data. Communications of the ACM, 39(11).

[12] Michailidis G, de Leeuw J (1998) The Gifi System of Descriptive Multivariate Analysis. Statistical Science, 13(4), 307-336.
[13] Wedel M., Kamakura W A (2000) Market segmentation: conceptual and methodological foundations. 2º edition. Kluwer Academic Publishers.
[14] Webb A (2002) Statistical Pattern Recognition, 2nd edition. John Wiley & Sons, Inc.
[15] Duda R O, Hart P E, Stork D G (2001) Pattern classification, 2nd edition. New York: John Wiley & Sons.
[16] Liao T W (2005) Clustering time series data – a survey. Pattern Recognition, 38, 1857–1874.
[17] Sahaa S, Bandyopadhyayb S (2012) Some connectivity based cluster validity indices. Applied Soft Computing, 12, 1555-1565.
[18] Rousseeuw P J (1987) Silhouettes: a graphical aid to the interpretation and validation of cluster analysis. Journal of Computational and Applied Mathematics, 20, 53-65.
[19] Cunha M R, Lambrecht B M, Pawlina G (2009) Determinants of Outstanding Mortgage Loan to Value Ratios: Evidence from the Netherlands. http://ssrn.com/abstract=1107822 . Accessed 16 May 2014.
[20] Georgarakos D, Lojschova A, Ward-Warmedinger M (2010) Mortgage Indebtedness and Household Financial Distress. European Central Bank, Working Paper N. 1156.
[21] Diaz-Serrano L (2004) Income volatility and residential mortgage deliquency: evidence from 12 EU countries. IZA Discussion Paper Series, No. 1396.
[22] Mann A, Mann R J, Staples S (2012) Debt, Bankruptcy, and the Life Course. http://ssrn.com/abstract=1492845. Accessed 04 March 2014.
[23] Salomon I, Ben-Akiva M E (1983) The use of lifestyle concept in travel demand models. Environment and Planning A, 15(5), 623–638.
[24] Warren E, Tyagi A W (2003) The Two-Income Trap: Why Middle-Class Mothers and Fathers Are Going Broke. New York: Basic Books.
[25] Keys B J (2009) The Credit Market Consequences of Job Displacement. National Poverty Center Working Paper Series 09-08.